\documentclass[10pt,twocolumn,letterpaper]{article}

\usepackage{wacv}              
\usepackage{booktabs}
\usepackage{caption}
\usepackage{array}
\usepackage{acronym}
\usepackage{comment}
\usepackage{graphicx} 
\usepackage{amsmath,amssymb}

\usepackage[ruled,vlined]{algorithm2e}

\definecolor{wacvblue}{rgb}{0.21,0.49,0.74}
\usepackage[pagebackref,breaklinks,colorlinks,allcolors=wacvblue]{hyperref}

\def\wacvPaperID{7} 
\def\confName{WACV}
\def\confYear{2026}

\title{LOST-3DSG: Lightweight Open-Vocabulary 3D Scene Graphs with Semantic Tracking in Dynamic Environments}

\author{Sara Micol Ferraina$^{*,1}$ \quad Michele Brienza$^{*,1}$\quad Francesco Argenziano$^1$ \quad Emanuele Musumeci$^1$ \\
Vincenzo Suriani$^1$ \quad Domenico D. Bloisi$^2$ \quad Daniele Nardi$^1$
\\ \\
$^1$Sapienza University of Rome, Rome, Italy\\
$^2$International University of Rome UNINT, Rome, Italy\\
$^*$Authors contributed equally\\
{\tt\small ferraina.1857726@studenti.uniroma1.it} \\
{\tt\small \{brienza, argenziano, musumeci, suriani, nardi\}@diag.uniroma1.it}\\
{\tt\small domenico.bloisi@unint.eu}
}

\acrodef{3DSGs}{3D Scene Graphs}
\acrodef{FMs}{Foundation Models}
\acrodef{VLMs}{Vision-Language Models}
\acrodef{LLMs}{Large Language Models}
\acrodef{PM}{Perception Module}
\acrodef{SUM}{Scene Update Module}
\acrodef{LSF}{Lost Similarity Function}

\begin{document}
\maketitle
\begin{abstract}
Tracking objects that move within dynamic environments is a core challenge in robotics. Recent research has advanced this topic significantly; however, many existing approaches remain inefficient due to their reliance on heavy foundation models. To address this limitation, we propose LOST-3DSG, a lightweight open-vocabulary 3D scene graph designed to track dynamic objects in real-world environments. Our method adopts a semantic approach to entity tracking based on word2vec and sentence embeddings, enabling an open-vocabulary representation while avoiding the necessity of storing dense CLIP visual features. As a result, LOST-3DSG achieves superior performance compared to approaches that rely on high-dimensional visual embeddings. We evaluate our method through qualitative and quantitative experiments conducted in a real 3D environment using a TIAGo robot. The results demonstrate the effectiveness and efficiency of LOST-3DSG in dynamic object tracking. Code and supplementary material are publicly available on the project website at https://lab-rococo-sapienza.github.io/lost-3dsg/.
\end{abstract}    
\section{Introduction}
\label{sec:intro}
The ability to reconstruct and represent the surrounding environment is a fundamental requirement for autonomous robots. This task is inherently challenging due to the need to recognize objects across different scales, handle duplicate instances, and remain robust to varying lighting conditions. The problem becomes even more complex in dynamic settings, where objects may change position or appearance over time. Without an accurate and continuously updated model of the world, a robot’s capacity to reason, plan, and act safely and effectively is severely constrained.\looseness-1

Humans naturally build internal representations of their environment and continuously refine them as the world evolves. This process involves tracking entities over time, understanding their motion, and maintaining semantic consistency despite changes. Replicating these capabilities in robots remains a significant challenge, particularly in real-world scenarios characterized by partial observability and frequent environmental changes.\looseness-1

Recent work on scene representation has increasingly adopted \ac{3DSGs}~\cite{armeni2019scenegraph, wald2020learning3d} as a flexible and expressive framework for modeling complex environments. By integrating geometric structure with semantic information, \ac{3DSGs} describe scenes as collections of object-centric nodes augmented with attributes and relational edges. This abstraction shifts scene understanding from low-level geometry to a structured, object-level representation that explicitly captures entities and their relationships.\looseness-1

Within this line of research, open-vocabulary \ac{3DSGs}~\cite{gu2024conceptgraphs, koch2024open3dsg, werby2024hierarchical, maggio2024clio} have gained significant attention. These approaches leverage large pre-trained \ac{FMs}, including \ac{VLMs} and CLIP~\cite{radford2021learning}, to encode object nodes with rich semantic representations that are not limited to a predefined taxonomy. As a result, robots can recognize and reason about previously unseen objects and concepts, substantially improving generalization in unstructured environments.\looseness-1

Despite these advantages, such expressiveness often comes at a significant computational and memory cost. Many existing methods rely on storing dense CLIP embeddings at the voxel or point level, producing large-scale semantic maps that are expensive to construct, update, and maintain~\cite{gu2024conceptgraphs, yan2025dynamic}. These costs are intensified in dynamic environments, where frequent updates are necessary to reflect changes in object pose and state. The repeated extraction, storage, and management of high-dimensional semantic features ultimately limit scalability and real-time applicability, highlighting the need for more efficient representations that preserve open-vocabulary reasoning while reducing computational overhead.\looseness-1

We propose \emph{LOST-3DSG}, a lightweight open-vocabulary 3D scene graph designed for dynamic environments. In contrast to existing approaches that store dense CLIP embeddings for objects in the scene, LOST-3DSG relies on low-cost \textit{word2vec}~\cite{mikolov2013efficient} embeddings derived from semantic attributes extracted using a VLM. These compact semantic representations enable robust tracking of dynamic objects over time by reasoning at the attribute level. Rather than depending solely on geometric consistency, the system determines whether an observation corresponds to a previously seen object or a new instance by matching its semantic attributes. For example, if an object previously identified as a \texttt{"red and brown, wooden and metal hammer"} reappears at a different location, it is more likely the same object that has moved rather than a newly observed one. By benchmarking our method against CLIP-based approaches and conducting an extensive ablation study, we demonstrate that our system can accurately track objects in the scene while maintaining a low computational footprint. To further validate the proposed approach, we deploy it in a real-world environment using a TIAGo robot\footnote{https://pal-robotics.com/robot/tiago/}. \looseness-1
We summarize our contributions as follows:
\begin{itemize}
\item \emph{LOST-3DSG}, a lightweight open-vocabulary 3D scene graph tailored for dynamic environments;
\item an efficient semantics-based tracking algorithm that updates the 3DSG as previously observed objects move or temporarily disappear;
\item an extensive experimental evaluation, including comparisons with CLIP-based methods, ablation studies, and real-world deployment on a TIAGo robot, demonstrating accurate object tracking with a limited computational footprint.
\end{itemize}

The remainder of this paper is organized as follows. Section~\ref{sec:rw} reviews related work on \ac{3DSGs} and scene representations for dynamic environments. Section~\ref{sec:methodology} describes the proposed approach in detail. Sections~\ref{sec:experimental_setup} and~\ref{sec:results} present the experimental setup and the obtained results, which are then discussed in Section~\ref{sec:discussion}. Finally, Section~\ref{sec:conclusions} draws the conclusions and outlines directions for future work.

\begin{figure*}
    \centering
    \includegraphics[width=\linewidth]{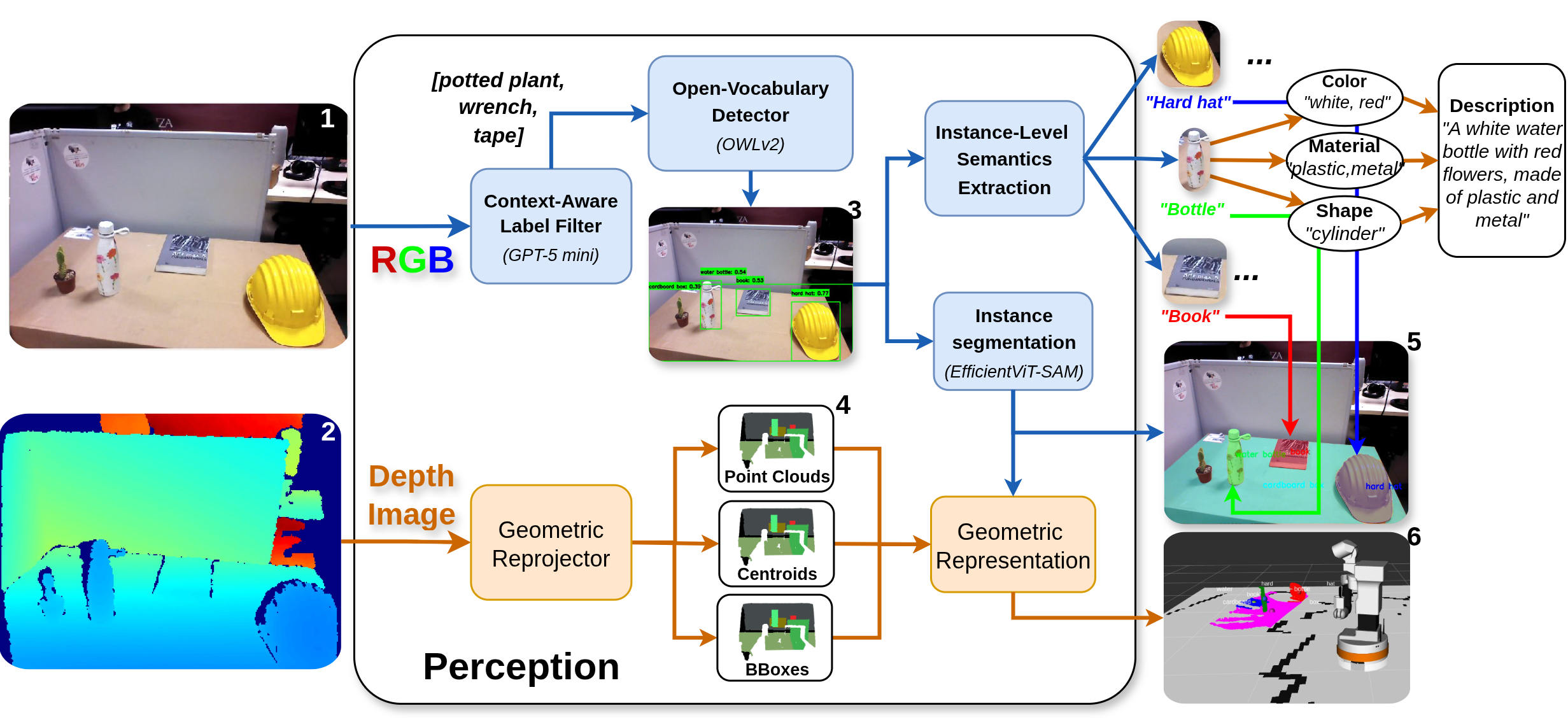}
    \caption{\textbf{Perception Module.} The current RGB frame (1) and the corresponding depth image (2) are processed to build the 3DSG of the scene. From the RGB image, open-vocabulary object labels are extracted using a VLM and then grounded in the image to detect the corresponding object bounding boxes on the camera plane (3). At the same time, the VLM is used to extract object-level semantic attributes, including label, color, material, and a fine-grained description. For each object instance, pixel-level segmentation masks are obtained using an object segmentation model and subsequently reprojected into 3D using depth information (5). In parallel, geometric primitives such as centroids and 2D bounding boxes are computed through geometric reprojection and used to estimate 3D bounding boxes (6).\looseness-1}
    \label{fig:fig1}
\end{figure*}
\section{Related Work}
\label{sec:rw}
\ \ \ \ \textbf{3D Scene Graphs} \ In recent years, \ac{3DSGs}~\cite{armeni2019scenegraph, wald2020learning3d, hughes2022hydra} have emerged as a prominent and widely adopted representation for modeling 3D environments. These representations describe a scene as a graph in which objects are encoded as nodes, while semantic, spatial, and functional relationships between them are captured through edges. This structured formulation provides a compact yet expressive way to model the entities present in an environment and to reason about their interactions. A key strength of \ac{3DSGs} lies in their flexibility. By tailoring node attributes and relational edges to the requirements of a given application, \ac{3DSGs} can support a broad range of downstream robotic tasks, including navigation~\cite{yin2024sg}, task planning~\cite{rana2023sayplan, liu2025delta, musumeci2025context}, manipulation~\cite{jiang2024roboexp, rotondi2025fungraph, honerkamp2024language}, and human-robot interaction~\cite{bartoli2025social, lim2025eventdrivenstorytellingmultiplelifelike}. Moreover, their graph-based structure makes them easily serializable in formats such as JSON, facilitating their integration into LLM-augmented applications~\cite{rana2023sayplan, chang2023dlite, gorlo2024trajectory}.\looseness-1

\textbf{Open-Vocabulary Scene Understanding} Recent advances in computer vision, together with the development of \ac{LLMs} and other \ac{FMs}, have enabled open-vocabulary scene understanding and parsing~\cite{Zhao_2017_ICCV}. These approaches allow scenes to be analyzed and objects to be detected without being constrained by a fixed, predefined taxonomy of labels, thereby improving generalization to novel and previously unseen concepts. A common strategy to achieve open-vocabulary understanding in 3D environments is to augment existing scene representations with semantic features extracted from pre-trained models such as CLIP~\cite{radford2021learning}. By associating CLIP embeddings with elements of a scene representation, whether voxels, points, primitives, or object nodes, it becomes possible to endow a wide range of mapping paradigms with rich, open-vocabulary semantics, including NeRF~\cite{engelmann2024opennerf}, Gaussian splatting~\cite{guo2024semantic}, point clouds~\cite{Peng_2023_CVPR}, and \ac{3DSGs}~\cite{koch2024open3dsg}.
In doing so, however, open-vocabulary \ac{3DSGs} such as ConceptGraphs~\cite{gu2024conceptgraphs} and DovSG~\cite{yan2025dynamic} require storing high-dimensional semantic features for each voxel in the scene, resulting in a queryable representation at the expense of a substantial memory footprint. In contrast, our method enables an open-vocabulary representation while requiring only a fraction of the computational resources used by CLIP-based approaches.\looseness-1

\textbf{Tracking in Dynamic Environments} \ Object tracking in dynamic environments has long been a fundamental challenge in robotics. Over the years, a wide range of approaches have been proposed, including filtering techniques such as Kalman filters~\cite{welch1995introduction}, probabilistic formulations~\cite{blackman2004multiple}, learning-based methods~\cite{ondruska2016deep}, and generative models~\cite{argenziano2025dynamic}. More recently, a growing body of work has focused on semantic tracking, where objects are tracked not only based on their motion but also according to their semantic properties and behaviors. In this direction, ~\citet{li2024beyond} propose tracking entities in video by jointly reasoning about both their trajectories, answering the question of where an object is, and the underlying semantic events, addressing what is happening. Similarly, ~\citet{zhang2020spatial} leverage deep convolutional features extracted from a pre-trained VGG~\cite{simonyan2014very} network to continuously track dynamic objects over time. SemTrack~\cite{wang2024semtrack} further advances this line of research by introducing the first large-scale dataset designed to train and evaluate models for semantic tracking in unconstrained environments. Inspired by these approaches, we design a semantic tracking algorithm that maintains object identity by reasoning over semantic attributes such as color and material, while updating object positions within the scene graph as the environment evolves. To the best of our knowledge, this is the first method to perform semantic object tracking directly within a 3DSG representation.\looseness-1

\section{Methodology}
\label{sec:methodology}
To construct a lightweight 3DSG from sensory observations, LOST-3DSG relies on two main components: the Perception Module and the Scene Update Module. The Perception Module processes raw sensor data to generate an initial 3DSG of the environment and, at the same time, extracts the semantic attributes associated with each object node. These attributes are later exploited for semantic tracking across observations. The Scene Update Module integrates newly acquired information into the previously constructed 3DSG, which serves as a persistent world anchor. Leveraging the proposed semantic tracking algorithm, this module detects when previously observed object instances reappear, move, or disappear as new observations are processed, and updates the 3DSG accordingly.\looseness-1

This section is organized as follows. Section~\ref{sec:methdology:pm} describes the Perception Module and its main components. Section~\ref{sec:metho:lost} details a similarity score based on objects semantic attributes. Finally, Section~\ref{sec:metho:sum} explains how the Scene Update Module uses this function to perform semantic tracking and to update the 3DSG over time.

\subsection{Perception Module}
\label{sec:methdology:pm}
The core objective of the \ac{PM} (Fig.~\ref{fig:fig1}) is to generate a 3DSG representation of the scene. Formally, we define a 3DSG as a hierarchical graph $\mathcal{G} = (\mathcal{V}, \mathcal{E})$, where each node $v \in \mathcal{V}$ represents an object in the environment. In terms of hierarchical structure, we consider three layers: the \texttt{room} layer, the \texttt{supporting objects} layer, and the \texttt{objects} layer. Edges are defined only between adjacent layers and encode \emph{belonging} relationships. This design choice reflects our focus on object-centric modeling, as the extraction of more complex semantic relations between nodes lies outside the scope of this work. Each node is enriched with a set of attributes that are essential for the proposed semantic tracking algorithm. In addition to an object \emph{label} $\ell$, we assign to each node its 3D \emph{bounding box} $b_{\text{3D}}\in\mathbb{R}^6$, \emph{color} $c$, \emph{material} $m$, and a short \emph{description} $d$ capturing fine-grained visual characteristics. This description supports instance-level discrimination between objects. In real-world environments, multiple objects may share the same label, material, and color while still differing in subtle visual details, such as surface patterns, wear, or scratches.\looseness-1

The \ac{PM} builds upon the EMPOWER framework~\cite{argenziano2024empower}, with several modifications tailored to continuous perception in dynamic environments. In particular, we extend EMPOWER with a streaming processing pipeline that incrementally handles sequential observations as the robot navigates through the environment. A first modification concerns open-set object labeling. In the original EMPOWER pipeline, a multi-role planner identifies task-relevant objects, which are then passed to \emph{YOLO-World}~\cite{cheng2024yolo} for detection. Since our objective is not task planning but scene understanding, we replace this component with a direct VLM query. Given an RGB observation $I_t$ at time $t$, the VLM produces a set of object labels
$
\mathcal{L}_t = \{ \ell_1, \ell_2, \dots, \ell_{N_t} \},
$
corresponding to the objects present in the scene. In our implementation, we use \emph{GPT-5-mini} as the VLM. The extracted labels are then passed to an open-vocabulary object detector to spatially ground the symbols in the image. Formally, the detector maps the image and label set to a collection of 2D bounding boxes
$
\mathcal{B}_t = \{ b_i \mid b_i = f_{\text{det}}(I_t, \ell_i) \},
$
where each bounding box $b_i$ localizes the corresponding object $\ell_i$. Unlike EMPOWER, which relies on \emph{YOLO-World}, we adopt \emph{OWLv2}~\cite{minderer2023scaling} due to its superior open-vocabulary detection performance. Each detected bounding box is then processed by \emph{EfficientViT-SAM}~\cite{zhang2024efficientvit} to obtain a pixel-level segmentation mask
$
m_i = f_{\text{seg}}(I_t, b_i),
$
which precisely delineates the spatial extent of the object in the image. In parallel, the cropped image region corresponding to each bounding box is analyzed by the VLM to extract a set of semantic attributes
$
a_i = \{ \ell, c, m, d \},
$
which are later used by the semantic tracking module. Using camera intrinsics $\mathbf{K}$, extrinsics $\mathbf{T}$, and the depth map $D_t$, each segmentation mask is reprojected into 3D space to label the point cloud:
\begin{equation}
\mathcal{P}_i = \pi^{-1}(m_i, D_t, \mathbf{K}, \mathbf{T}),
\end{equation}
where $\pi^{-1}$ denotes the back-projection operation. This step associates each 3D point with its corresponding object instance.
Finally, the downstream components of EMPOWER are employed to extract object nodes from the segmented point cloud and to infer hierarchical relations between them, resulting in the final 3DSG representation of the environment produced by the \ac{PM}.

\subsection{Lost Similarity Function}
\label{sec:metho:lost}
To associate object observations over time, we introduce the \emph{\ac{LSF}}, a composite metric designed to quantify the likelihood that two observations, namely a current detection and an existing node in the 3DSG, correspond to the same physical instance. The function integrates multiple complementary cues into a single similarity score, capturing both semantic and appearance-based consistency. The \ac{LSF} is defined as the weighted combination of four terms:
\begin{itemize}
\item \textit{Semantic similarity} ($s_{\ell}$): computed using pre-trained \emph{word2vec} embeddings to measure the cosine similarity between object labels. This term captures semantic relatedness between categories, allowing the tracker to associate objects even when their labels slightly differ but remain conceptually related.
\item \textit{Chromatic similarity} ($s_c$): computed by converting object colors into RGB space and measuring their normalized Euclidean distance. The similarity is defined as
\begin{equation}
    s_c = 1 - \frac{\lVert \mathbf{rgb}_1 - \mathbf{rgb}_2 \rVert_2}{\sqrt{3}},
\end{equation}
where normalization by $\sqrt{3}$ ensures that $s_c \in [0,1]$.

\item \textit{Material similarity} ($s_m$): analogous to semantic similarity, this term uses \emph{word2vec} embeddings to compare the material attributes of objects, enabling robust matching across observations despite minor variations in appearance.

\item \textit{Description similarity} ($s_d$): computed using a Sentence Transformer (specifically, OpenAI's \emph{text-embedding-3-small} model) to embed fine-grained textual descriptions of objects. This term captures detailed visual characteristics and supports instance-level discrimination between objects that share the same label, color, and material.
\end{itemize}

The final score is obtained as a weighted linear combination of these components:
\begin{equation}
LSF(o_1,o_2) = \alpha s_{\ell} + \beta s_c + \gamma s_m + \delta s_d,
\end{equation}
where $\alpha$, $\beta$, $\gamma$, and $\delta$ control the relative contribution of each similarity term, and $\alpha +\beta+\gamma+\delta=1$. These weights can be adjusted to balance semantic and appearance cues depending on the characteristics of the environment and the desired tracking behavior. In our experiment we set $\alpha=0.15$, $\beta=0.30$, $\gamma=0.15$ and $\delta=0.40$ as we empirically noticed the best results were achieved with this combination.


\begin{figure}
    \centering
    \includegraphics[width=\linewidth]{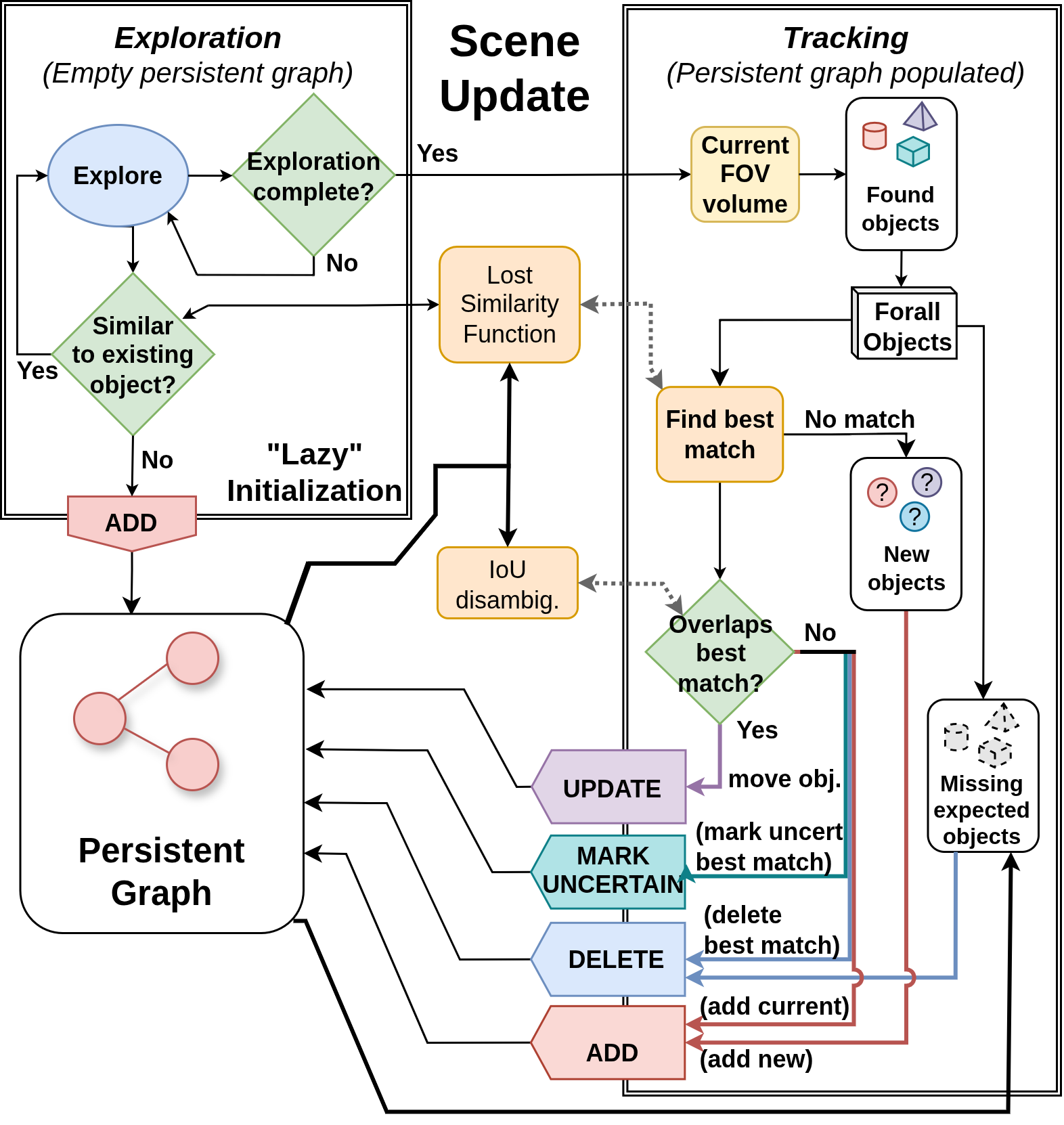}
    \caption{\textbf{Scene Update Module}. During the \textit{exploration} phase, the Persistent 3DSG is incrementally populated using LSF-based disambiguation. Once exploration ends, the system switches to the \textit{tracking} phase, where objects in the current FOV are matched by semantic similarity and spatial consistency: moved objects are updated, new objects are added, and missing objects are removed or marked as uncertain.}
    \vspace{-0.5cm}
    \label{fig:sum}
\end{figure}

\subsection{Scene Update Module}
\label{sec:metho:sum}
The \ac{SUM} (Fig.~\ref{fig:sum}) is formalized in the \texttt{scene\_update} procedure described in Algorithm~\ref{algo:scene_update}. To associate a current detection with an object already present in the 3DSG, the system computes the \ac{LSF} between the detection and all persistent objects. The object with the highest similarity score is selected as the best candidate match, provided that the score exceeds a minimum threshold $\tau$. If no such candidate exists, the detection is treated as a previously unseen object. The \ac{SUM} operates in two modes, controlled by the exploration flag $e$: \textit{exploration} and \textit{tracking}, and the behavior of the system differs significantly between these modes.\looseness-1

\textbf{Exploration mode} When the exploration flag $e$ is active, the system focuses on populating the 3DSG by incrementally discovering objects as the robot navigates the environment. For each detection, the system attempts to associate it with an existing persistent object. If no valid association is found, a new object node is spawned and added to the set of persistent objects. If a match exists, the corresponding bounding box is updated, but no object is removed or marked as uncertain. This design choice reflects the assumption that, during exploration, all observations contribute to building a complete catalog of the environment. As a result, object removal and identity conflict resolution are intentionally disabled to avoid prematurely discarding valid objects.\looseness-1

\textbf{Tracking mode} \ Once exploration is complete, the system switches to tracking mode by setting $e = \texttt{false}$. In this phase, the scene manager actively maintains consistency between the 3DSG and the current observations. For each detection $d \in \mathcal{D}$, the algorithm proceeds as follows. First, the best matching persistent object $p$ is retrieved using the LOST similarity function. If no match is found, a new object is spawned and marked as observed in the current frame. Otherwise, the validity of the association is evaluated based on spatial consistency. If the association is deemed valid, the bounding box of the persistent object is updated and the object is marked as seen in the current frame. If the association is invalid, indicating a semantic match but inconsistent spatial evidence, the system resolves the ambiguity by removing the object from its previous location and marking it as uncertain. A new object instance is then spawned at the newly observed position. The removed object is stored in the \texttt{uncertain\_objects} set, which preserves nodes affected by identity conflicts and allows potential recovery if future observations resolve the ambiguity.\looseness-1

\textbf{Graph maintenance and cleanup} \ At the end of each update cycle in tracking mode, the system performs a cleanup step based on the robot’s point-of-view (POV). The visible volume $V$ is computed from the current camera pose, and persistent objects that were not observed in the current frame but lie within the POV volume are pruned. Similarly, uncertain objects that remain unobserved despite being within the visible region are also removed. This mechanism ensures that objects are only removed when they should have been visible but were not detected, thereby reducing false deletions due to occlusions or limited sensor coverage.

\begin{algorithm}[t]
\small
\DontPrintSemicolon
\SetAlgoSkip{smallskip}
\SetInd{0.3em}{0.6em}
\setlength{\algomargin}{0.5em}
\SetNlSkip{0.5em}

\SetKwInOut{Input}{Input}
\SetKwInOut{Output}{Output}

\SetKwFunction{BBox}{FindBBox}
\SetKwFunction{Match}{FindBestMatch}
\SetKwFunction{Valid}{IsValidAssociation}
\SetKwFunction{NewObj}{SpawnObject}
\SetKwFunction{Upd}{UpdateBBox}
\SetKwFunction{Seen}{MarkSeen}
\SetKwFunction{Flag}{MarkUncertainAndRemove}
\SetKwFunction{POV}{ComputePOVVolume}
\SetKwFunction{PruneP}{PrunePersistentObjects}
\SetKwFunction{PruneU}{PruneUncertainObjects}
\SetKwProg{Fn}{Function}{:}{end}

\Input{
detections $\mathcal{D}$, persistent objects $\mathcal{P}$, current scene $\mathcal{S}$, exploration flag $e$
}

\Output{
updated persistent objects $\mathcal{P}$ and uncertain objects $\mathcal{U}$
}

\Fn{\texttt{scene\_update}}{
    $\mathcal{S}_{\text{seen}} \leftarrow \emptyset$\;

    \ForEach{$d \in \mathcal{D}$}{
        $b \leftarrow \BBox(d)$\;
        \lIf{$b = \emptyset$}{\textbf{continue}}

        $p \leftarrow \Match(d, \mathcal{P})$\tcp*[r]{LSF}
        \uIf{$p = \emptyset$}{
            \NewObj($d, b, \mathcal{S}_{\text{seen}}$)\;
        }
        \Else{
            \uIf{$e$}{
                \Upd($p, b$)\;
            }
            \uElseIf{\Valid($p, b$)}{
                \Upd($p, b$)\;
                \Seen($p, \mathcal{S}_{\text{seen}}$)\;
            }
            \Else{
                \Flag($p$)\;
                \NewObj($d, b, \mathcal{S}_{\text{seen}}$)\;
            }
        }
    }

    \If{$\lnot e$}{
        $V \leftarrow \POV(\mathcal{S})$\;
        \PruneP($V, \mathcal{S}_{\text{seen}}$)\;
        \PruneU($V$)\;
    }
}
\caption{Scene Update Algorithm}
\label{algo:scene_update}
\end{algorithm}
    
\section{Experimental Setup}
\label{sec:experimental_setup}
This section describes the experimental setup used to evaluate the proposed approach in a controlled laboratory environment. Experiments were conducted using a TIAGo robot equipped with ROS 2 Humble and operating in a real-world indoor setting. The robot is provided with a prior map of the environment and, during the exploration phase, continuously localizes itself while detecting and estimating the spatial positions of nearby objects. All observations are subsequently projected into the global map reference frame.~\looseness-1

\begin{figure}
    \centering
    \includegraphics[width=\linewidth]{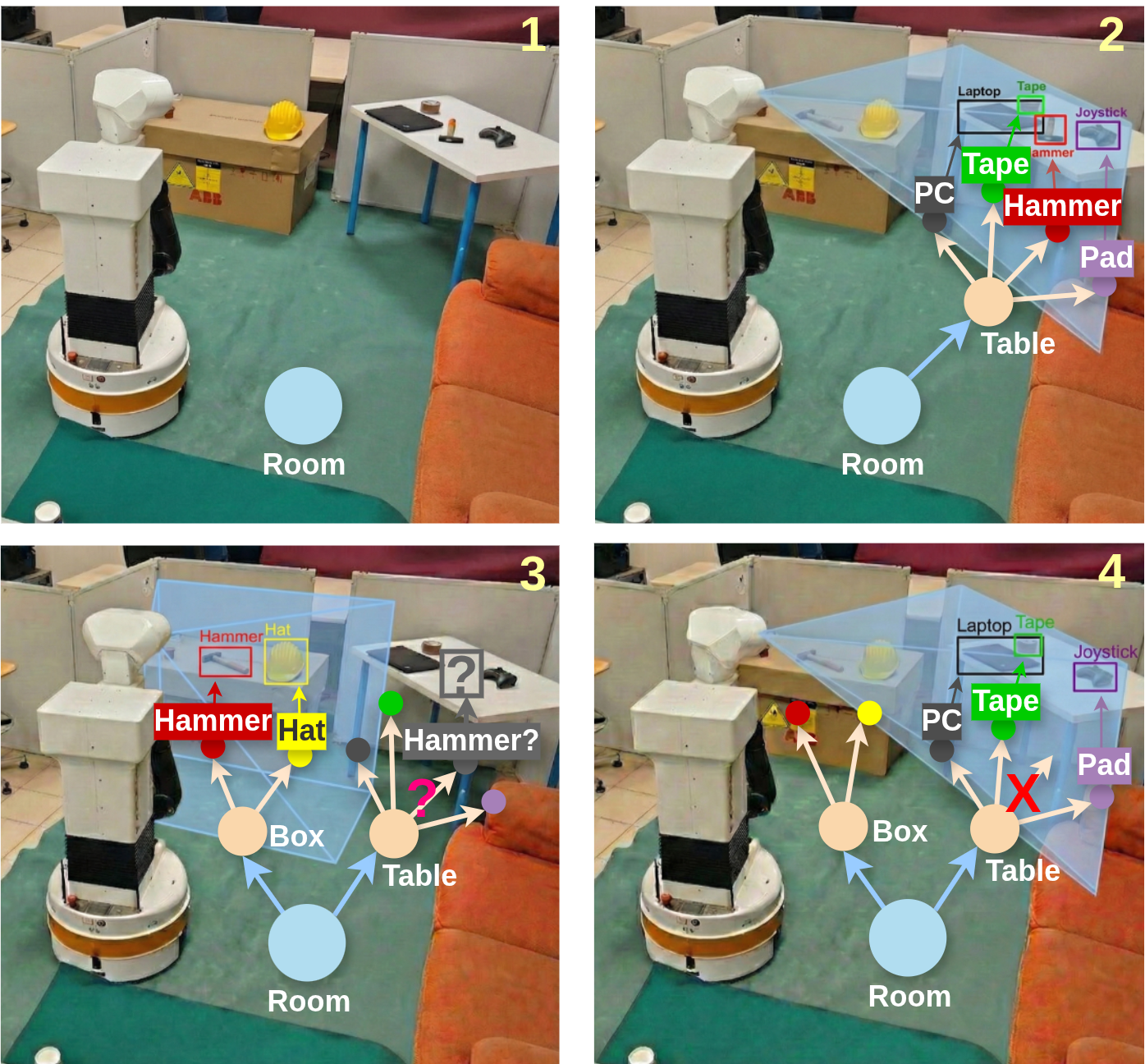}
    \caption{\textbf{Execution instance.} The agent observes a household environment (1). During the \textit{exploration} phase, a laptop, a tape roll, a hammer, and a gamepad are added to the 3DSG (2). The hammer is then moved to the brown surface on the left. When the hammer is observed at its new location, the previous instance is marked as uncertain (3), and once the original location is revisited and the object is confirmed absent, the instance is removed from the 3DSG (4).}
    \label{fig:placeholder}
\end{figure}

To systematically assess the behavior of the system under increasing levels of complexity, a set of experimental scenarios was designed, featuring multiple tables and objects distributed throughout the environment. In total, three scenes were created, ranging from simple static configurations to more challenging dynamic setups. The evaluated scenarios are summarized as follows, with increasing level of complexity:
\begin{enumerate}
\item \textbf{Level ($\star$):} the robot observes a scene containing three initially static objects that, over time, move, change position, and eventually disappear from the environment.
\item \textbf{Level ($\star\ \star$):} the environment contains a substantially larger number of objects. While the robot explores the scene from multiple viewpoints, several objects change position and some are removed entirely. This level is more challenging due to the increased object density combined with partial observations, object updates, and disappearances that occur outside the robot’s field of view.~\looseness-1
\item \textbf{Level ($\star\star\star$):} the robot operates in a smaller but highly dynamic environment. Objects undergo numerous position updates, with frequent changes occurring while they are unobserved. This level stress-tests the system’s semantic tracking capabilities, requiring it to consistently associate object identities and infer updated locations across time and viewpoints.
\end{enumerate}
An example of execution is given in Fig.~\ref{fig:placeholder}.

\begin{table}[t]
\centering
\caption{\ac{SUM} performance at different Level of Complexity (LoC).}
\label{tab:q1}
\begin{tabular}{c c c c c c c c}
LoC &
Objects &
Detections &
Deletions &
Updates \\
\hline
\hline
$\star$        & 3 & 3/3 & 3/3 & 3/3 \\
$\star\ \star$   & 21 & 20/21 & 2/3 & 1/1 \\
$\star\star\star$ & 9 & 2/3 & 2/3 & 13/14 \\
\bottomrule
\end{tabular}
\end{table}

\section{Results}
\label{sec:results}
In evaluating our system, we aim to answer the following research questions:
\textbf{Q1.} How effective is the \ac{SUM} at tracking objects using semantic information alone?
\textbf{Q2.} What is the relative contribution of each component of the \ac{LSF} to accurate object identification and association?
\textbf{Q3.} To what extent does the proposed approach reduce memory consumption compared to CLIP-based methods that store dense semantic features?~\looseness-1

Answers to Q1 are shown in Table~\ref{tab:q1}, that highlights both the system performance at different level of complexity, as discussed in Sec.~\ref{sec:results}. The system demonstrates strong robustness in handling object updates across different difficulty levels. In the medium scenario, the main source of complexity arises from the substantially larger number of objects in the environment. Several objects change position and some are deleted while outside the robot’s field of view. As a result, the primary challenge lies in correctly retrieving the appropriate object instance through similarity matching, where ambiguities in semantic associations can lead to errors. The hard scenario is particularly demanding due to the high number of object updates. In this setting, many objects frequently change position, often without being directly observed. This configuration effectively stress-tests the system’s semantic tracking capabilities, requiring persistent object identity maintenance and accurate graph updates over time. Despite these challenges, the system is able to consistently associate updated observations with previously seen objects and correctly reflect their new positions in the 3DSG.

We perform an ablation study to answer Q2, and the results are shown in Table~\ref{tab:q2}, averaged across the three levels of complexity. Detections are not reported since the \ac{LSF} does not affect that component. The full LSF achieves the highest update stability and balanced deletion behavior. When only description similarity is retained, update performance remains relatively strong, confirming that fine-grained textual descriptions are effective at maintaining instance identity over time. However, deletions degrade noticeably in this setting, indicating that descriptions alone are insufficient to robustly decide when an object should be removed. Using only semantic labels leads to poor performance across both metrics, particularly for deletions, due to the susceptibility of label-based matching to hallucinations and category ambiguity. This confirms that labels alone are not reliable for long-term object persistence. Removing material and chromatic similarity does not affect deletions, but significantly degrades update accuracy, supporting the interpretation that low-level appearance cues are primarily used to disambiguate visually similar instances during temporal updates rather than for object lifecycle decisions. Overall, the results highlight the complementary roles of description, semantic, and appearance cues within the \ac{LSF}. 

Finally, we answer Q3 by reporting a comparison of the memory footprint of our system against CLIP-based approaches that rely on storing dense semantic features. We consider the lowest-dimensional CLIP configuration, namely \emph{ViT-B/32}, which produces embeddings of 512 floating-point elements. Assuming 16-bit precision, each embedding requires $512 \times 2\text{ B} = 1024\text{ B} \approx 1\text{ KB}$. In our most complex experimental scene, 21 objects are present and the environment is represented at a voxel resolution of 25\,mm per side, resulting in a total number of voxels equal to $ N_{\text{voxels}} = 626\,140 $ for the scene. Storing a CLIP embedding for each voxel would therefore require a total memory of $M_{\text{CLIP}} = N_{\text{voxels}} \times 1024\text{ B} \approx 641 \text{ MB}$. In contrast, our approach stores semantic information only at the object level. For each object, we require at most 157\,B: 12\,B for the two bounding box extents, assuming 16-bit floating-point coordinates, up to 100\,B for the textual description, and up to 15\,B each for the material, color, and label attributes, assuming UTF-8 encoding with one byte per character. For the entire scene, this results in a total memory usage of $M_{\text{LOST-3DSG}} = 21 \times 157\text{ B} = 3297\text{ B}.$ This corresponds to slightly more than 3\,KB to represent the scene, while preserving open-vocabulary expressiveness and enabling efficient tracking of dynamic objects.

\begin{table}[t]
\centering
\caption{Ablation study on the \ac{LSF} components, averaged through the various LoCs.}
\label{tab:q2}
\setlength{\tabcolsep}{6pt}
\begin{tabular}{c c c c c c c c}
\ac{LSF} &
Deletions &
Updates \\
\hline
\hline
Full         & \textbf{0.778} & \textbf{0.944} \\
$s_d$, $s_m$, $s_c$  & 0.556 &  0.889 \\
$s_\ell$, $s_m$, $s_c$ & 0.667 & 0.667 \\
$s_\ell$, $s_d$ & 0.667 & 0.778 \\
$s_d$ & 0.444 & 0.833\\
$s_l$ & 0.333 & 0.556 \\
\bottomrule
\end{tabular}
\end{table}
\section{Discussion}
\label{sec:discussion}
The results indicate that lightweight, attribute-level semantic representations can support object tracking in dynamic environments without relying on dense visual embeddings. The \ac{SUM} maintains object identity across time and varying scene complexity, correctly handling object updates and disappearances under partial observability, although errors still occur in the presence of semantic ambiguities. The ablation study shows that fine-grained descriptions play a central role in instance discrimination, while chromatic and material cues contribute to stable temporal updates, and label-only similarity is insufficient for long-term persistence. Also, storing semantics exclusively at the object level yields substantial memory savings compared to CLIP-based voxel representations. However, this efficiency comes with a modest performance cost in challenging scenarios where richer visual features could improve disambiguation. In particular, variability in VLM-generated descriptions can lead to identity fragmentation when similarity scores fall below the matching threshold, and the absence of temporal aggregation makes the system sensitive to noisy or inconsistent semantic estimates. Conservative graph update policies may also introduce temporary object duplication when semantic and spatial evidence disagree. Despite these limitations, the results suggest that the proposed approach offers a promising trade-off between efficiency and tracking accuracy, and that incorporating more robust object profiling, explicit temporal aggregation, and adaptive similarity modeling could further narrow the performance gap while preserving scalability.
\section{Conclusion}
\label{sec:conclusions}

In this paper, we introduced LOST-3DSG, a lightweight open-vocabulary 3DSG designed to support semantic object tracking in dynamic environments without relying on dense Foundation Models embeddings. By operating on compact, attribute-level representations, the proposed approach significantly reduces memory and computational overhead while still enabling reliable instance association over time. Through real-world experiments, we demonstrated that simple semantic cues, when combined in the proposed manner, are often sufficient to maintain consistent object identities across motion, partial observability, and viewpoint changes. Although the system is intentionally minimal, the results indicate that heavyweight visual embeddings are not always required for effective open-vocabulary scene understanding. This work is intended to establish foundational insights rather than provide a complete solution. Several directions remain open for future research, including more robust temporal aggregation of semantic descriptions, improved similarity modeling, and closer integration with downstream reasoning and planning modules. We hope this work encourages further exploration of lightweight and scalable alternatives to dense semantic mapping approaches, and serves as a starting point for future research on long-term, open-world 3DSG-based scene understanding.

\section*{Acknowledgements}
{
This work has been carried out while Michele Brienza, Francesco Argenziano and Emanuele Musumeci were enrolled in the Italian National Doctorate on Artificial Intelligence run by Sapienza University of Rome. Michele Brienza is founded by the European Union - Next Generation EU, Mission I.4.1 Borse PNRR Pubblica Amministrazione (Missione 4) Component 1 CUP B53C23003540006. We acknowledge partial financial support from PNRR MUR project PE0000013-FAIR.
}

{
    \small
    \bibliographystyle{ieeenat_fullname}
    \bibliography{main}
}

\end{document}